\pdfoutput=1

\documentclass[11pt]{article}

\usepackage[]{acl}

\usepackage{times}
\usepackage{latexsym}

\usepackage[T1]{fontenc}

\usepackage[utf8]{inputenc}

\usepackage{microtype}

\usepackage{inconsolata}
\usepackage{subfigure}
\usepackage{graphicx}
\usepackage{amssymb}
\usepackage{multirow}
\usepackage{pifont}
\usepackage{array}
\usepackage{enumitem}
\usepackage{bm}
\usepackage{amstext}
\usepackage{diagbox}
\usepackage{tabularx}
\usepackage{booktabs}
\usepackage{amsmath}
\setenumerate[1]{itemsep=0pt,partopsep=0pt,parsep=\parskip,topsep=0pt}
\setitemize[1]{itemsep=0pt,partopsep=0pt,parsep=\parskip,topsep=0pt}
\setdescription{itemsep=0pt,partopsep=0pt,parsep=\parskip,topsep=0pt}
\title{PVGRU: Generating Diverse and Relevant Dialogue Responses via Pseudo-Variational Mechanism}

\author{Yongkang Liu\textsuperscript{\rm 1,2,3}, Shi Feng\textsuperscript{\rm 1}, Daling Wang\textsuperscript{\rm 1}, Yifei Zhang\textsuperscript{\rm 1}, Hinrich Schütze\textsuperscript{\rm 2,3} \\
\textsuperscript{\rm 1} Northeastern University, China\\
\textsuperscript{\rm 2} Center for Information and Language Processing, LMU Munich\\
\textsuperscript{\rm 3} Munich Center for Machine Learning (MCML), LMU Munich\\
\texttt{misonsky@163.com, \{fengshi,wangdaling,zhangyifei\}@cse.neu.edu.cn}\\
}

\begin{document}
\maketitle
\begin{abstract}
We investigate response generation for multi-turn dialogue in generative-based chatbots. Existing generative models based on RNNs (Recurrent Neural Networks) usually employ the last hidden state to summarize the sequences, which makes models unable to capture the subtle variability observed in different dialogues and cannot distinguish the differences between dialogues that are similar in composition. In this paper, we propose a \textbf{P}seudo-\textbf{V}ariational \textbf{G}ated \textbf{R}ecurrent \textbf{U}nit (PVGRU) component without posterior knowledge 
through introducing a recurrent summarizing variable into the GRU, which can aggregate the accumulated distribution variations of subsequences.
PVGRU can perceive the subtle semantic variability through summarizing variables that are optimized by the devised distribution consistency and reconstruction objectives. In addition, we build a \textbf{P}seudo-\textbf{V}ariational \textbf{H}ierarchical \textbf{D}ialogue (PVHD) model based on PVGRU. Experimental results demonstrate that PVGRU can broadly improve the diversity and relevance of responses on two benchmark datasets.
\end{abstract}
\section{Introduction}
\begin{figure*}[t]
\centering
\includegraphics[width=0.90\textwidth]{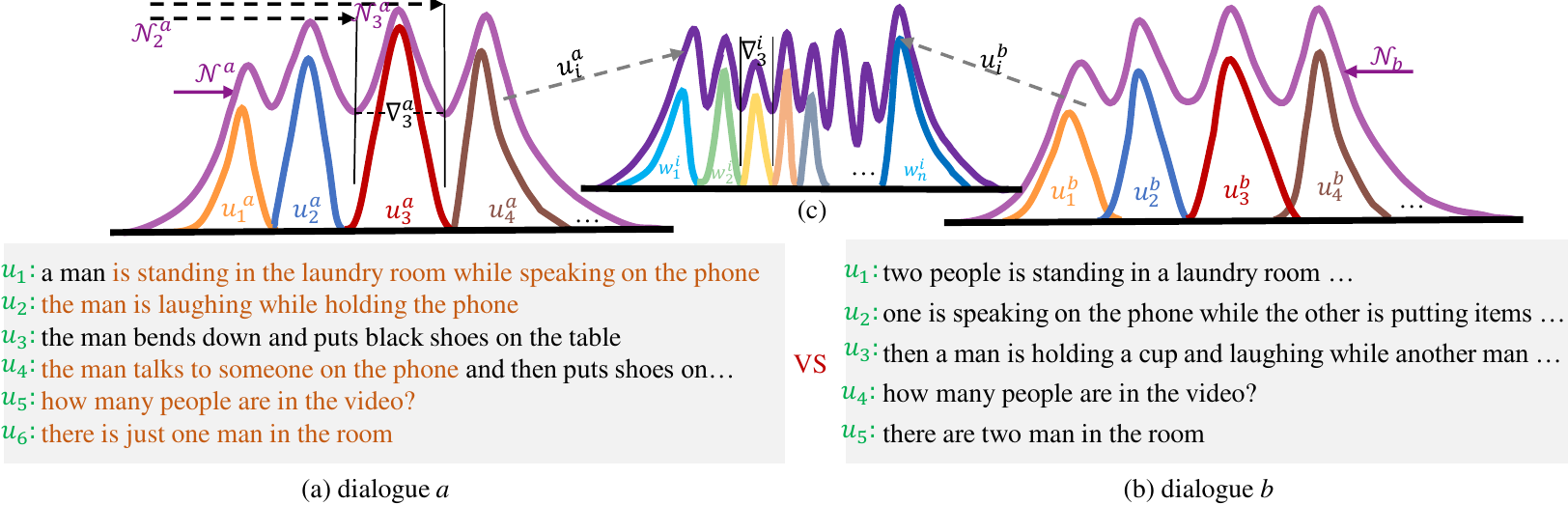}
\caption{Overview of dialogue variability. (a) and (b) represent two dialogues \textit{a} and \textit{b} from DSTC7-AVSD. $\mathcal{N}^a$ and $\mathcal{N}^b$ represent the distribution composition of dialogue \textit{a} and \textit{b} on utterance level, respectively. $\mathcal{N}_t^a$ represents the distribution of dialogue \textit{a} at time step \textit{t} on utterance level. (c) stands for the distribution composition of an utterance. $u_i^a$ and $u_i^b$ represent the \textit{i}-th utterance of dialogue \textit{a} and \textit{b}, respectively. $w_j^i$ stands for the \textit{j}-th word of the \textit{i}-th utterance. $\bigtriangledown_3^a$ denotes the distribution variations caused by $u_3$ to $\mathcal{N}^a$ and $\bigtriangledown_3^i$ denotes the variations caused by token $w_3^i$ to the distribution of $u_i$. The utterances marked in brown in dialogue \textit{a} indicate that there is a similar expression in dialogue \textit{b}.}
\label{example}
\end{figure*}

Complex grammatical rules exist in the high variability text data~\cite{gormley2015elasticsearch,chung2015recurrent,nie2022cross}, especially dialogue. As shown in Figure~\ref{example}, two utterances with just one/two words different may have different semantics, such as utterance $u_6$ of dialogue \textit{a} vs. $u_5$ of dialogue \textit{b}; 
On the other hand, two dialogues with lots of semantically similar utterances may express quite different context meanings, such as \textit{a} vs. \textit{b}. 
These variabilities lead to multiple mappings between dialogue context and response, which occurs in response causing \textit{one-to-many} issue and in context resulting in \textit{many-to-one} problem. We can observe that the distribution of dialogue contexts (i.e., $\mathcal{N}^a$ and $\mathcal{N}^b$) is composed of the distribution of utterances, and the distribution of each utterance consists of the distribution of words (i.e., Figure~\ref{example} (c)). How to model the word-level and utterance-level variation in dialogue plays an important role in improving the quality of responses.

A line of existing researches~\cite{henderson2014word,shang2015neural,serban2016building,luo2018auto} employ RNNs (Recurrent Neural Networks) to model dialogue context. However, researchers perceive that it is inappropriate to employ RNNs to directly model this kind of variability observed in dialogue corpora~\cite{chung2015recurrent}. This is because the internal transition structure of the RNNs is entirely deterministic, which can not effectively model the randomness or variability in dialogue contexts~\cite{chung2015recurrent}.

Variational mechanism has demonstrated attractive merits in modeling variability from both theoretical and practical perspectives~\cite{kingma2013auto}. Methods based on variational mechanism~\cite{serban2016building,gu2018dialogwae,khan2020adversarial,sun2021generating} introduce latent variable into RNNs
to model \textit{one-to-many} and \textit{many-to-one} phenomena in dialogue. Although these approaches achieve promising results, they still have defects. First, these methods face the dilemma that latent variables may vanish because of the posterior collapse issue~\cite{zhao2017learning,zhao2018unsupervised,shi2020dispersed}. Variational mechanism can work only when latent variables with intractable posterior distributions exist~\cite{kingma2013auto}. Second, the sampled latent variables may not correctly reflect the semantics of the dialogue context due to the one-to-many and many-to-one phenomena observed in dialogue~\cite{sun2021generating}. Third, the posterior knowledge is employed in training while prior knowledge used in inference, which causes an inconsistency problem between training and inference~\cite{shang2015neural,zhao2017learning,shi2020dispersed}.

To tackle these problems, we propose a \textbf{P}seudo-\textbf{V}ariational \textbf{G}ated \textbf{R}ecurrent \textbf{U}nit (PVGRU) component based on pseudo-variational mechanism. PVGRU is based on GRU by introducing a recurrent summarizing variable, which can aggregate the accumulated distribution variations of subsequences. The methods based on PVGRU can perceive the subtle semantic differences between different sequences.
First, the pseudo-variational mechanism adopts the idea of latent variables but does not adopts posterior mechanism~\cite{serban2017hierarchical,zhao2017learning,park2018hierarchical,sun2021generating}. Therefore, PVGRU does not suffer from the posterior collapse issue~\cite{zhao2017learning,zhao2018unsupervised,shi2020dispersed}. Second, we design the consistency and reconstruction objectives to optimize the recurrent summarizing variable in PVGRU, 
which ensure that the recurrent variable can reflect the semantic of dialogue context from word- and utterance-level, respectively. The consistency objective makes the distribution of the incremental information consistent with the corresponding input at each time step. For instance in Figure~\ref{example}, the distribution of $u_3^a$ is consistent with $\bigtriangledown_3^a=\mathcal{N}_3^a-\mathcal{N}_2^a$ and the distribution of $w_3^i$ is consistent with $\bigtriangledown_3^i$. The reconstruction objective demands the summarizing variable to be able to reconstruct the sequence information. For example, we can reconstruct the subsequence information before time step 3 from distribution $\mathcal{N}_3^a$. Third, we guarantees the consistency between training and inference since we do not employ posterior knowledge when optimizing summarizing variable.

In addition, we build a \textbf{P}seudo-\textbf{V}ariational \textbf{H}ierarchical \textbf{D}ialogue model (PVHD) based on PVGRU to model the word-level and utterance-level variation. To summarize, we make the following contributions:
\begin{itemize}[leftmargin=*]
    \item We analyze the reasons for \textit{one-to-many} and {\textit{many-to-one}} issues from high variability of dialogue corpus and propose PVGRU with recurrent summarizing variable to model the variability of dialogue sequences.
    \item We propose to optimize recurrent summarizing variable using consistency and reconstruction objective, which guarantees that the summarizing variable can reflect the semantics of the dialogue context and maintain the consistency between training and inference processes. 
    \item We propose PVHD model based on PVGRU, which significantly outperforms strong baselines with RNN and Transformer architectures on two benchmark datasets. The code including baselines for comparison is avaliable on Github\footnote{https://github.com/misonsky/PVHD}.
\end{itemize}
\section{RELATED WORK}
\subsection{Dialogue Generation}
As an important task in Natural Language Processing, dialogue generation systems aim to generate fluent and informative responses based on the dialogue context~\cite{ke2018generating}. Early dialogue generation models~\cite{henderson2014word,shang2015neural,luo2018auto} usually adopt the simple \textit{seq2seq}~\cite{sutskever2014sequence} framework to model the relationship between dialogue context and response in the manner of machine translation. However, the vanilla seq2seq structure tends to generate dull and generic responses. To generate informative responses, hierarchical structures~\cite{serban2016building,song2021bob,liu2022mulzdg} and pre-training techniques~\cite{radford2019language,lewis2019bart,zhang2020dialogpt} are employed to capture the hierarchical dependencies of dialogue context. The results of these methods do not meet expectations~\cite{wei2019neural}.

The main reason is that there are one-to-many and many-to-one relationships between dialogue context and responses.
Modeling the multi-mapping relationship is crucial for improving the quality of the dialog generation. In this paper, we propose a PVGRU component by introducing recurrent summarizing variables into GRU, which can model the varieties of dialogue context.
\subsection{Variational Mechanism}
Variational mechanisms enable efficient working in directed probabilistic models when latent variables with intractable posterior distributions existing~\cite{kingma2013auto}. Variational mechanisms can learn the latent relationship between dialogue context and responses by introducing latent variables. Most existing methods~\cite{serban2017hierarchical,zhao2017learning,bao2020plato} based on variational mechanisms employ prior to approximate true posterior probability. These methods not only encounter the problem of posterior collapse issue but also the problem of inconsistency between training and inference~\cite{zhao2018unsupervised,shi2020dispersed}. In this paper, we employ consistency and reconstruction objectives to optimize the summarizing variable different from variational mechanism, which can model the multi-mapping phenomena in dialogues.
\section{Preliminary}
In this paper, we employ GRU (Gated Recurrent Unit)~\cite{cho2014learning} as the implementation of recurrent neural network (RNN). The reset gate $r_t$ is computed by:
\begin{equation}
\small
r_t = \sigma(\bm{W}_r\bm{x}_{t} + \bm{U}_r\bm{h}_{t-1})
\end{equation}
where $\sigma$ is the logistic sigmoid function. $\bm{x}_t$ represents the input at time step $t$ and $\bm{h}_{t-1}$ denotes the hidden state at time step $t$-1. $\bm{W}_r$ and $\bm{U}_r$ are parameter matrices which are learned. Similarly, the updated gate $z_t$
is defined as:
\begin{equation}
\small
z_t = \sigma(\bm{W}_z\bm{x}_{t} + \bm{U}_z\bm{h}_{t-1})
\end{equation}
The hidden state $\bm{h}_t$ at the time step $t$ is then computed by:
\begin{equation}
\small
\bm{h}_t = z_t\bm{h}_{t-1} + (1-z_t)\bm{\tilde{h}}_t
\end{equation}
\begin{equation}
\small
\bm{\tilde{h}}_t = \phi(\bm{W}x_t + \bm{U}(r_t \odot \bm{h}_{t-1}))
\end{equation}
where $\phi(\cdot)$ is the tanh function, $\bm{W}$ and $\bm{U}$ are weight matrices which are learned. GRU is considered as a classic implementation of RNN, which is widely employed in generative tasks.
\begin{figure*}[t]
\centering 
\includegraphics[width=0.90\textwidth]{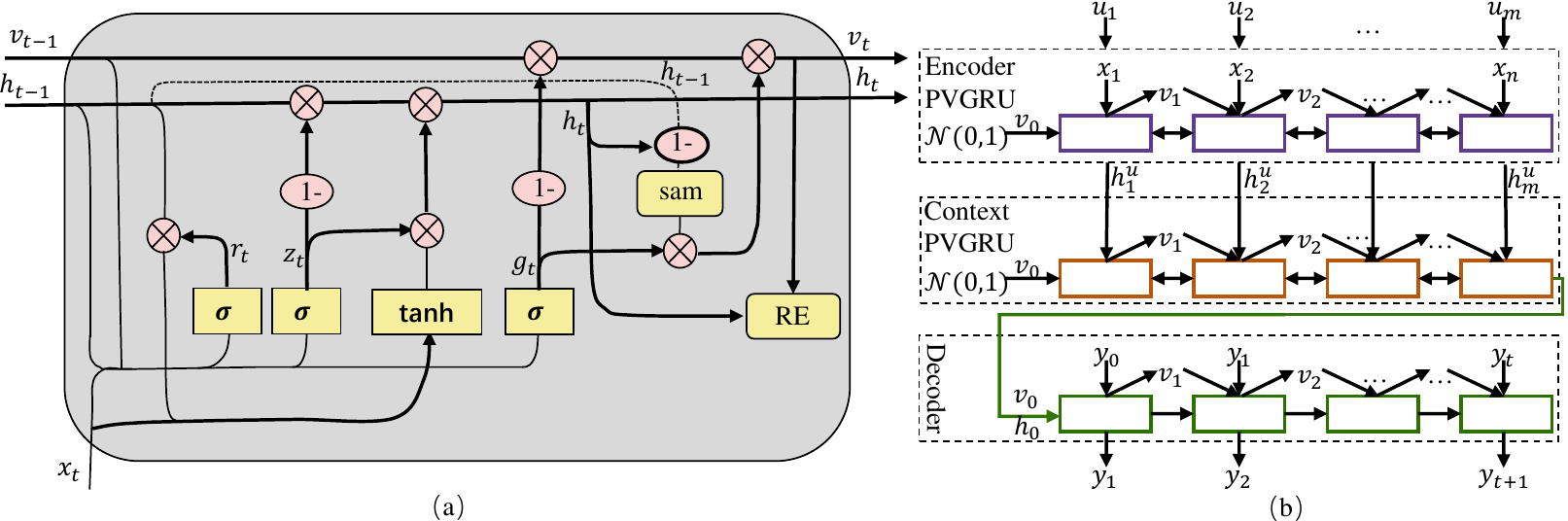}
\caption{Overview of PVHD based on PVGRU. (a) is the overview of PVGRU, where RE stands for refactoring process and the "sam" represents sampling process. (b) is graphical representation of the PVHD.}
\label{pvgru}
\end{figure*}
\begin{figure}[t]
\centering 
\includegraphics[width=0.90\columnwidth]{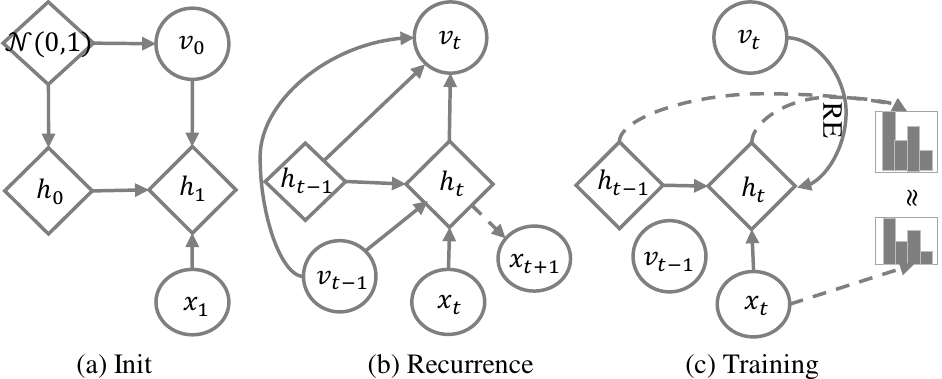}
\caption{Schematic diagram of each operation of PVGRU autoregression.}
\label{schematic}
\end{figure}
\section{Methodology}
\subsection{Pseudo-variational Gated Recurrent Unit}
As shown in Figure~\ref{example}, it is difficult to distinguish the semantics of similar dialogue contexts only relying on the last hidden state representations. The internal transition structure of RNNs is deterministic, which can not model variability observed in dialogues and tends to generate dull and generic responses. Drawing the inspiration from variational recurrent neural network (VRNN)~\cite{chung2015recurrent}, our proposed PVGRU explicitly models the variability through introducing a recurrent summarizing variable, which can capture the variations of dialogue context. VRNN based on variational mechanism employs latent variables paying attention to the variety between different words. Different from VRNN, PVGRU maintains a summarizing variable unit that can summarize the accumulated variations of the sequence.

As shown in Figure~\ref{pvgru} (a), PVGRU introduces a recurrent summarizing variable $\bm{v}$ based on GRU. The recurrent summarizing variable $\bm{v}$ is obtained based on the incremental information of hidden state $\bm{h}$ and the previous state of summarizing variable. Specially, the summarizing variable $\bm{v}_0$ is initialized with standard Gaussian distribution (i.e., Figure~\ref{schematic} (a)). We assume the input is $x_t$ at the time step $t$, the reset gate $r_t$ is rewrited as:
\begin{equation}
\small
r_t = \sigma(\bm{W}_r\bm{x}_{t} + \bm{U}_r\bm{h}_{t-1} + \bm{V}_r\bm{v}_{t-1})
\end{equation}
where $\bm{W}_r$, $\bm{U}_r$ and $\bm{V}_r$ are parameter matrices, and $v_{t-1}$ is the previous summarizing variable state. 
Similarly, the update gate $z_t$ is computed by:
\begin{equation}
\small
z_t = \sigma(\bm{W}_z\bm{x}_{t} + \bm{U}_z\bm{h}_{t-1} + \bm{V}_z\bm{v}_{t-1})
\end{equation}
We introduce a gate $g_t$ for summarizing variable factor, which is defined as follows:
\begin{equation}
\small
\bm{g}_t = \sigma(\bm{W}_gx_t + \bm{U}_g\bm{h}_{t-1} + \bm{V}_g\bm{v}_{t-1})
\end{equation}
The updated gate of summarizing factor controls how much information from the previous variable will carry over to the current summarizing variable state. Under the effect of $g_t$, the $\bm{\tilde{h}}_t$ follows the equation:
\begin{equation}
\small
\bm{\tilde{h}}_t = \phi(\bm{W}x_t + \bm{U}(r_t \odot \bm{h}_{t-1}) + \bm{V}(\bm{g}_t \odot \bm{v}_{t-1}))
\end{equation}
Then the PVGRU updates its hidden state $\bm{h}_t$ using the same recurrence equation as GRU. The summarizing variable $v_{t}$ at the time step $t$ is defined as:
\begin{equation}
\small
\bm{\tilde{v}}_{t} \sim \mathcal{N}(\mu_t,\sigma_t), [\mu_t,\sigma_t] = \varphi(\bm{h}_{t} - \bm{h_{t-1}})
\end{equation}
where $\varphi(\cdot)$ represents a nonlinear neural network approximator and $\bm{\tilde{v}}_{t}$ denotes the variations between time $t$ and time $t-1$. The variations across subsequent up to time $t$ is defined as:
\begin{equation}
\small
\bm{v}_{t} = \bm{g}_t \odot \bm{\tilde{v}}_{t} + (1-\bm{g}_t) \odot \bm{v}_{t-1}
\end{equation}
Figure~\ref{schematic} (b) demonstrates the schematic diagram of the recurrent process of PVGRU described above. We can observe that PVGRU does not adopt posterior knowledge, which can guarantee the consistency between training and inference.
\subsection{Optimization Summarizing Variable}
Based on but different from traditional variational mechanism, we design the consistency and reconstruction objectives to optimize the summarizing variable. The consistency objective ensures that the distribution of the information increment of hidden state at each time step is consistent with the input. For example, we will keep the distribution of information increment $h_t-h_{t-1}$ at time $t$ consistent with $x_t$. The consistency objective function at time step $t$ is denoted as:
\begin{equation}
\small
\begin{split}
\ell_{c}^t &= KL(p(\bm{x}_t)||p(\bm{h}_t-\bm{h}_{t-1})) \\
         &= KL(p(\bm{x}_t)||\tilde{\bm{v}_t})
\end{split}
\end{equation}
where \textit{KL}($\cdot$) represents Kullback-Leibler divergence~\cite{barz2018detecting} and \textit{p}($\cdot$) represents the distribution of the vector. We employ "sam" to represent this process of distribution sampling in Figure~\ref{pvgru} (a).

The reconstruction optimization objective ensures that the summarizing variable can correctly reflect the semantic of the dialogue context from the whole perspective, which requires PVGRU reconstructs the sequence information from the accumulated distribution variable. The reconstruction loss at time step $t$ is described as:
\begin{equation}
\small
\ell_r^t(\bm{v}_t,\bm{h}_t)=\left\{
\begin{array}{ll}
\frac{1}{2}(f(\bm{v}_t)-\bm{h}_t), & {|\bm{v}_t-\bm{h}_t| \leq \delta} \\
\delta |f(\bm{v}_t)-\bm{h}_t|-\frac{1}{2} \delta^2, & {|\bm{v}_t-\bm{h}_t| > \delta}
\end{array} \right.
\end{equation}
where $f(\cdot)$ stands for decoder using MLP, $\delta$ is a hyperparameter and $|\cdot|$ represents the absolute value. We employ "RE" to represent the reconstruction process in Figure~\ref{pvgru} (a).
Figure~\ref{schematic} (c) demonstrates the schematic diagram of optimizing summarizing variable. Reconstruction and consistency objectives ensure that summarizing variable can correctly reflect the semantics of the dialogue context.
\subsection{Hierarchical Pseudo-variational Model} 
As shown in Figure~\ref{example}, the dialogues contain word-level and sentence-level variability. We follow previous studies~\cite{serban2016building,serban2017hierarchical,huang2021attention} using hierarchical structure to model dialogue context. Figure~\ref{pvgru} (b) shows the structure of PVHD we proposed. PVHD mainly consists of three modules: (i) Encoder PVGRU; (ii) Context PVGRU; (iii) Decoder PVGRU. The encoder PVGRU is responsible for capturing the word-level variabilities and mapping utterances$\{\bm{u}_1,\bm{u}_2,...,\bm{u}_m\}$ to utterance vectors $\{\bm{h}_1^u,\bm{h}_2^u,...,\bm{h}_m^u\}$. At the same time, $\bm{v}_t$ records the accumulated distribution variations of the subsequence at time step \textit{t}. The context PVGRU takes charge of capturing the utterance-level variabilities. The last hidden state of the context PVGRU represents a summary of the dialogue. The last summarizing variable state of the context PVGRU stands for the distribution of dialogue. The decoder PVGRU takes the last states of context PVGRU and produces a probability distribution over the tokens in the response $\{y_1,y_2,...,y_n\}$.
The generation process of training and inference can be formally described as:
\begin{equation}
\small
p(\bm{y}_{\leq T},\bm{v}_{\leq n}) = \prod_{t=1}^{n}p(\bm{y}_t | \bm{y}_{< t}, \bm{v}_{< t})
\end{equation}
The log-likelihood loss of predicting reponse is formalized as:
\begin{equation}
\small
\ell_{ll}^t= \text{log} p(\bm{y}_t |\bm{y}_{<t},\bm{v}_{<t})
\end{equation}
The total loss can be written as:
\begin{equation}
\small
\ell_{total} = E\sum_{t=1}^T(\ell_{ll}^t+\ell_r^t+\ell_c^t)
\end{equation}

\begin{table*}[h]
\small
\centering
\begin{tabular}{lcccccccc}
\toprule
Models & Datasets & Types & PPL & BLEU-1/2 & Rouge-L & Dist-1 & Dist-2 & Embed A/E/G \\
\multirow{4}{*}{seq2seq}
  &\multirow{2}{*}{Daily}
  & GRU & 132.55& 27.78/22.59& 35.36& 12.18 & 47.69& 79.40/80.02/63.53\\
  & & PVGRU &130.80& 28.33/22.48& 36.55& 14.41&48.22& 80.77/81.26/63.96\\
  \cline{2-9}
  &\multirow{2}{*}{DSTC7}
  & GRU &112.89& 25.52/15.29& 26.34& 4.34&22.31& 79.31/84.40/60.25\\
  & & PVGRU &111.27& 26.66/17.18& 27.72& 5.77&24.68& 80.56/85.65/60.48\\
\midrule
\multirow{4}{*}{HRED}
  &\multirow{2}{*}{Daily}
  & GRU &127.66& 28.90/23.52& 34.63& 13.00&45.55& 79.53/81.77/63.31\\
  & & PVGRU &111.31& 32.19/25.42& 35.28& 15.33&49.93&  81.77/83.89/63.84 \\
  \cline{2-9}
  &\multirow{2}{*}{DSTC7}
  & GRU &115.72& 27.30/17.86& 29.51& 5.12&24.63& 79.18/84.78/61.71\\
  & & PVGRU &110.25& 29.87/20.03& 31.87& 6.54&31.77& 81.87/86.68/61.91\\
\midrule
\multirow{4}{*}{HRAN} 
  &\multirow{2}{*}{Daily}
  & GRU &121.63& 30.36/20.01& 35.68 &12.66&43.77& 80.42/84.56/63.44\\
  & & PVGRU &120.77& 30.97/23.76& 36.52& 13.76&44.86& 81.05/85.58/63.35\\
  \cline{2-9}
  &\multirow{2}{*}{DSTC7}
  & GRU &111.66& 27.74/17.88& 30.68& 4.64&17.68& 80.31/82.33/62.70\\
  & & PVGRU &110.75& 29.58/19.68& 32.34& 5.33&19.62& 81.86/85.34/63.34\\
\midrule
\multirow{4}{*}{CSG} 
  &\multirow{2}{*}{Daily}
  & GRU &122.75& 28.89/24.55& 36.74 &11.11&40.39& 79.65/83.36/63.29\\
  & & PVGRU &122.12& 30.04/26.67& 38.39& 13.21&42.44& 80.83/84.55/65.95\\
  \cline{2-9}
  &\multirow{2}{*}{DSTC7}
  & GRU &111.27& 27.62/18.24& 28.32 &3.07&12.13& 79.55/82.19/62.27\\
  & & PVGRU &110.82& 29.74/20.55& 31.02& 5.13&15.44& 80.53/84.91/63.18\\
\bottomrule
\end{tabular}
\caption{Performance comparison of models based on GRU and PVGRU on on test set of DailyDialog (Daily) and DSTC7-AVSD (DSTC7). All values are multiplied by 100.}
\label{tab:pvgru}
\end{table*}
\begin{figure}[t]
\centering
\includegraphics[width=0.90\columnwidth]{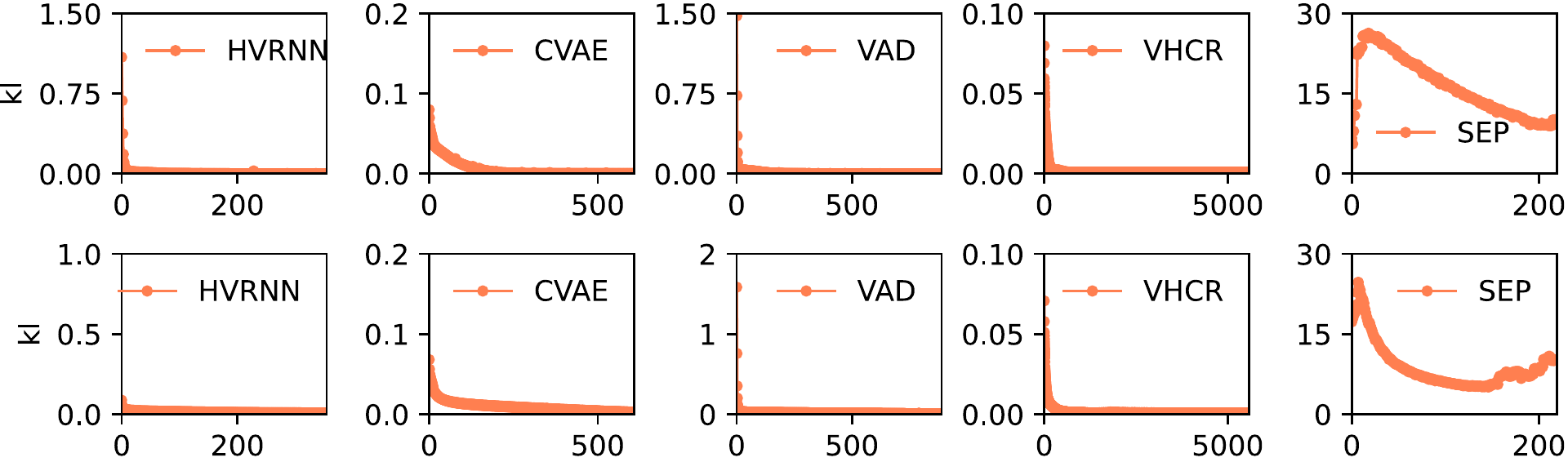}
\caption{Kullback-Leibler loss variation trend graph on DailyDialog (up) and DSTC7-AVSD (down). The abscissa represents the number of training iterations. KL represents the Kullback-Leibler loss term.}
\label{klloss}
\end{figure}
\section{Experiments}
For descriptions of the datasets, please refer to the Appendix~\ref{app:datasets}.
Please refer to Appendix~\ref{app:impl} for implementation details. In Appendix~\ref{app:ablation} we show the ablation results of two objective functions, showing the effectiveness of the objective functions. In order to evaluate the effectiveness of experimental results, we performed a significance test in Appendix~\ref{app:sign}. We can observe that the \textit{p-values} of PVHD are less than 0.05 compared with other models. In addition, we present case studies in Appendix~\ref{app:case} and discuss model limitations in Appendix~\ref{app:lim}, respectively.
\subsection{Baselines}
\label{app:baselines}
The automatic evaluation metrics is employed to verify the generality of PVGRU, we select the following RNN-based dialogue generation models as baselines:
\textbf{seq2seq}: sequence-to-sequence model GRU-based with attention mechanisms~\cite{bahdanau2014neural}.
\textbf{HRED}: hierarchical recurrent encoder-decoder on recurrent neural network~\cite{serban2016building} for dialogue generation.
\textbf{HRAN}: hierarchical recurrent neural network dialogue generation model based on attentiom mechanism~\cite{xing2018hierarchical}.
\textbf{CSG}: hierarchical recurrent neural network model using static attention for context-sensitive generation of dialogue responses~\cite{zhang2018context}.

To evaluate the performance of the PVHD, we choose dialogue generation model based on variational mechanism as baselines:
\textbf{HVRNN}: VRNN (Variational Recurrent Neural Network)~\cite{chung2015recurrent} is a recurrent version of the VAE. We combine VRNN~\cite{chung2015recurrent} and HRED~\cite{serban2016building} to construct the HVRNN.
\textbf{CVAE}: hierarchical dialogue generation model based on conditional variational autoencoders~\cite{zhao2017learning}. We implement CVAE with bag-of-word loss and KL annealing technique.
\textbf{VAD}: hierarchical dialogue generation model introducing a series of latent variables~\cite{du2018variational}.
\textbf{VHCR}: hierarchical dialogue generation model using global and local latent variables~\cite{park2018hierarchical}.
\textbf{SepaCVAE}: self-separated conditional variational autoencoder introducing group information to regularize the latent variables~\cite{sun2021generating}.
\textbf{SVT}: sequential variational transformer augmenting deocder with a sequence of fine-grained latent variables~\cite{lin2020variational}.
\textbf{GVT}: global variational transformer modeling the discourse-level diversity with a global latent variable~\cite{lin2020variational}.
\textbf{PLATO}: dialogue generation based on transformer with discrete latent variable~\cite{bao2020plato}. Different from original implementation, we do not use knowledge on the DSTC7-AVSD.
\textbf{DialogVED}: a pre-trained latent variable encoder-decoder model for dialog response generation~\cite{chen2022dialogved}. We initialize the model with the large version of DialogVED.
\begin{table*}[h]
\small
\centering
\begin{tabular}{lcccccccc}
\toprule
Datasets & Backbone & Models  & PPL & BLEU-1/2 & Rouge-L & Dist-1 &Dist-2 & Embed A/E/G \\
\midrule
\multirow{10}{*}{Daily}
& \multirow{4}{*}{Transformer}
& SVT &114.54& 27.89/21.26& 28.87& 11.94&44.03&  77.67/83.39/60.14 \\ 
 &  & GVT &115.05& 25.54/18.46& 26.87& 12.43&45.43&  75.90/83.16/56.42 \\
 &  & PLATO &\textbf{110.68}& 30.77/24.46& 33.95& 13.41&47.67&  79.15/\textbf{84.15}/60.09 \\ 
 &  & DialogVED &112.87& \underline{31.22}/\underline{24.96}& 33.16& 12.94&45.44& 78.36/83.73/60.25 \\ \cline{2-9}
& \multirow{6}{*}{RNN}
& HVRNN & 124.94& 31.03/23.99& 34.83& \underline{14.32}&49.47&  79.55/83.75/62.03\\
  & & CVAE & 126.38& 26.34/20.43& \underline{35.83}& 13.55&49.18&  79.70/83.45/63.26\\
 & & VAD & 134.06& 30.32/24.34& \textbf{36.63}& 13.85&46.20&  \underline{80.97}/\underline{84.09}/\textbf{63.87}\\
  & & VHCR & 115.83& 29.80/24.35& 34.45& 13.66&\underline{49.50}&  79.01/81.27/62.35\\
  & & SepaCVAE & 111.33& 25.31/22.41& 33.21& 12.08&36.56&  80.26/81.81/63.51\\ 
  & & PVHD &\underline{111.31}& \textbf{32.19}/\textbf{25.42}& 35.28& \textbf{15.33}&\textbf{49.93}&  \textbf{81.77}/83.89/\underline{63.84} \\ \hline
\midrule
\multirow{10}{*}{DSTC7}
& \multirow{4}{*}{Transformer}
& SVT &116.58& 25.34/14.28& 25.47& 3.67&15.75& 78.88/82.87/56.87 \\ 
  & & GVT &115.33& 27.62/15.76& 26.71& 3.14&17.49& 77.56/84.07/57.46 \\
  & & PLATO &\textbf{108.88}& \textbf{30.16}/18.58& \underline{30.69}& 6.22&29.39& 80.05/85.71/58.22 \\
  & & DialogVED &112.09& 28.89/13.69& 29.22& \underline{6.39}&26.78& 79.36/85.73/60.25 \\ \cline{2-9}
& \multirow{6}{*}{RNN}
  & HVRNN & 111.55& 26.71/18.12& 29.44& 5.52&21.23& 79.76/86.51/60.11\\
  & & CVAE & 112.40& 26.47/16.37& 28.85& 5.35&26.01& \underline{80.96}/\textbf{86.88}/\underline{60.68}\\
  & & VAD & 122.37& 26.87/\textbf{20.26}& 27.07& 6.00&\underline{30.46}& 79.24/86.41/58.37\\
  & & VHCR & 123.81& 26.63/15.81& 28.21& 5.64&29.83& 79.71/86.65/57.56\\
  & & SepaCVAE & 128.47& 26.59/18.94& 26.04& 5.53&28.50& 78.85/86.31/59.06\\
  & & PVHD &\underline{110.25}& \underline{29.87}/\underline{20.03}& \textbf{31.87}& \textbf{6.54}&\textbf{31.77}& \textbf{81.07}/\underline{86.68}/\textbf{61.91}\\ \hline
\bottomrule
\end{tabular}
\caption{Performance comparison of PVHD and other models based on variational mechanism. \textbf{Bold} indicates the best result, and \underline{underline} indicates the second best result. The first and second groups of models belong to the Transformer-based models and RNN-based models, respectively.}
\label{tab:PVHD}
\end{table*}
\subsection{Automatic \& Human Evaluation}
Please refer to Appendix~\ref{app:metrics} and Appendix~\ref{app:human} for details of automatic evaluation metrics. Some differences from previous works are emphasized here. We employ improved versions of BLEU and ROUGE-L, which can better correlate n-gram overlap with human judgment by weighting the relevant n-gram compared with original BLEU~\cite{chen2014systematic}. Although using the improved versions of BLEU and ROUGE-L will result in lower literal values on the corresponding metrics, this does not affect the fairness of the comparison. We adopt the implementation of distinct-1/2 metrics following previous study~\cite{bahuleyan2018variational}. The source code for the evaluation method can be found on the anonymous GitHub.
\begin{table}[ht]
\small
\centering
\begin{tabular}{p{1.2cm}p{.6cm}p{.6cm}p{.6cm}|p{.6cm}p{.6cm}p{.6cm}}
\toprule
  \multirow{3}{*}{Models} & \multicolumn{6}{c}{Datasets}  \\ \cline{2-7}
  & \multicolumn{3}{c|}{DailyDialog} &  \multicolumn{3}{c}{DSTC7-AVSD} \\
  & D & R & F & D & R & F \\ \hline
  SVT &0.920 & 0.795 & 1.752 &0.973 & 1.115 & 1.271 \\
  GVT &0.950 & 0.769 & 1.780 &0.950 & 1.046 & 1.361 \\
  PLATO &\underline{1.110} & 0.847 & 1.783 &\underline{1.087} & \underline{1.437} & \underline{1.742} \\
  DialogVED &1.090 & \textbf{0.856} & 1.830 &1.010 & 1.372 & 1.540 \\ \hline
  HVRNN & 1.000 & 0.780 & \textbf{1.850} & 1.041 & 1.415 & \textbf{1.785}  \\
  CVAE & 1.080 & 0.765 & 1.450 & 1.025 & 1.085 & 1.100 \\
  VAD & 1.015 & 0.854 & 1.235 & 0.990 & 1.215 & 1.400  \\
  VHCR & 0.895 & 0.835 & 1.570 & 0.975 & 1.250 & 1.600 \\
  SepaCVAE &1.020 & 0.695 & 1.230 & 1.040 & 0.715 & 0.810 \\
  PVHD & \textbf{1.114} & \underline{0.855} & \underline{1.840} & \textbf{1.145} & \textbf{1.445} & 1.520 \\ \hline
\bottomrule
\end{tabular}
\caption{Human evaluation results on test set. D, R, F represent diversity, relevance and fluency, respectively.}
\label{tab:human_evaluation}
\end{table}
\begin{figure}[t]
\centering
\includegraphics[width=0.90\columnwidth]{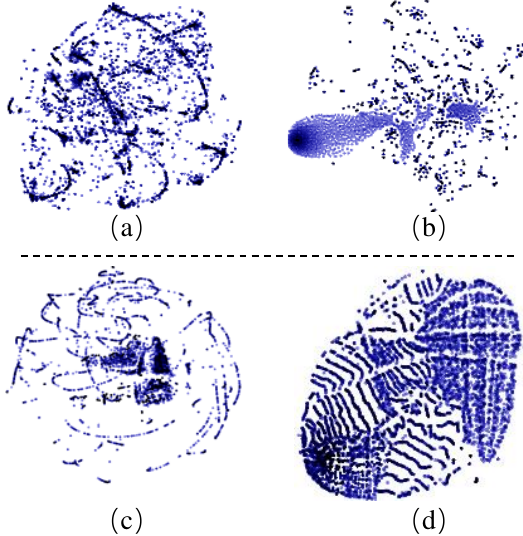}
\caption{t-SNE visualization of the summarizing variable on word-level ((a) and (c)) and utterance-level (b) and (d) on DailyDialog (up) and DSTC7-AVSD (down).}
\label{visiual}
\end{figure}

\subsection{Generality of PVGRU}
Table~\ref{tab:pvgru} reports the automatic evaluation performance comparison of the models using GRU and PVGRU. We can observe that the performance of the models based on PVGRU is higher than that based on GRU. Specifically, on DailyDialog dataset, the average performance of models based on PVGRU is 0.63\% to 16.35\% higher on PPL, 1.40\% to 1.92\% higher on BLEU-1, 1.08\% to 2.02\% higher on Rouge-L, 1.10\% to 2.33\% higher on Dist-1 and 1.36\% to 1.62\% higher on average embedding compared with models based on GRU. On DSTC7-AVSD dataset, the performance of models based on PVGRU is 0.45\% to 5.47\% higher on PPL, 1.14\% to 2.57\% higher on BLEU-1, 1.38\% to 2.7\% higher on Rouge-L, 0.69\% to 2.06\% higher on Dist-1 and 0.69\% to 2.69\% higher on average embedding compared with models based on GRU. The results demonstrate that PVGRU can be widely used to sequence generation models based on RNN. The internal transition structure of GRU is entirely deterministic. Compared with GRU, PVGRU introduces a recurrent summarizing variable, which records the accumulated distribution variations of sequences. The recurrent summarizing variable brings randomness to the internal transition structure of PVGRU, which makes model perceive the subtle semantic variability.
\subsection{Automatic Evaluation Results \& Analysis}
Table~\ref{tab:PVHD} reports the results of automatic evaluation of PVHD and other baselines on DailyDialog and DSTC7-AVSD datasets. Compared to RNN-based baselines based on variational mechanism, PVHD enjoys an advantage in performance. On DailyDialog datasets, the performance of PVHD is 1.16\% higher on BLEU-1, 0.45\% higher on Rouge-L, 1.01\% higher on Dist-1 and 2.22\% higher on average embedding compared to HVRNN. As compared to the classic variational mechanism models CVAE, VAD and VHCR, PVHD has a advantage of 0.02\% to 22.75\% on PPL, 1.87\% to 6.88\% higher on BLEU-1, 1.48\% to 3.25\% higher on Dist-1, 0.43\% to 13.37\% higher on Dist-2 and 0.80\% to 2.76\% higher on average embedding. We can observe similar results on DSTC7-AVSD. PVHD enjoys the advantage of 1.3\% to 18.22\% on PPL, 3.00\% to 3.40\% higher on BLEU-1, 0.54\% to 1.19\% higher on Dist-1, 1.31\% to 5.76\% higher on Dist-2 and 0.11\% to 2.22\% higher on average embedding compared with these classic variational mechanism models.

The main reason for the unimpressive performance of RNN-based baselines is that these models suffer from latent variables vanishing observed in experiments.
As shown in Figure~\ref{klloss}, the Kullback-Leibler term of these models losses close to zero means that variational posterior distribution closely matches the prior for a subset of latent variables, indicating that failure of the variational mechanism~\cite{lucas2019don}. The performance of SepaCVAE is unimpressive. In fact, the performance of SepaCVAE depends on the quality of context grouping (referring to dialogue augmentation in original paper~\cite{sun2021generating}). SepaCVAE will degenerate to CVAE model if context grouping fails to work well, and even which will introduce wrong grouping noise information resulting in degrade performance. As shown in Figure~\ref{klloss}, the Kullback-Leibler term of SepaCVAE losses is at a high level, which demonstrates that the prior for a subset of latent variables cannot approximate variational posterior distribution.

Compared with Transformer-based baselines, PVHD still enjoys an advantage on most metrics, especially the distinct metric. GVT introduces latent variables between the whole dialogue history and response, which faces the problem of latent variables vanishing. SVT introduces a sequence of latent variables into the decoder to model the diversity of responses. But it is debatable whether latent variables will destroy the fragile sequence perception ability of the transformer, which will greatly reduce the quality of the responses. Training the transformer from scratch instead of using a pretrained model is another reason for the inferior performance of SVT and GVT. Compared to DialogVED and PLATO, PVHD achieves the best performance on most metrics. The main reason is that pseudo-variational approaches do not depend on posteriors distribution avoiding optimization problems and the recurrent summarizing variable can model the diversity of sequences. Overall, PVHD has the most obvious advantages in diversity, which demonstrates the effectiveness of the recurrent summarizing variable.

Although transformers are popular for generation task, our research is still meritorious. First, transformer models usually require pre-training on large-scale corpus while RNN-based models usually do not have such limitations. It is debatable whether transformer models training from scratch under conditions where pre-training language models are unavaliable can achieve the desired performance if downstream task does not have enough corpus. Second, the parameter amount of the RNN-based model is usually smaller than that of the transformer-based model. The parameter sizes of PVHD on the DailyDialog and DSTC7-AVSD are 29M and 21M, respectively. The number of parameters for PLATO and DialogVED is 132M and 1143M on two datasets, respectively. Compared to PLATO and DialogVED,  the average number of parameters of PVHD is 5.28x and 45.72x smaller, respectively.
\subsection{Human Evaluation Results \& Analysis}
We conduct human evaluation to further confirm the effectiveness of the PVHD. To evaluate the consistency of the results assessed by annotators, we employ Pearson’s correlation coefficient~\cite{sedgwick2012pearson}. This coefficient is 0.35 on diversity, 0.65 on relevance, and 0.75 on fluency, with $p$< 0.0001 and below 0.001, which demonstrates high correlation and agreement. The results of the human evaluation are shown in Table~\ref{tab:human_evaluation}. Compared to RNN-based baselines, PVHD has a significant advantage in relevance and diversity. Specifically, PVHD enjoys the advantage of 11.40\% on diversity and 16.00\% on relevance compared to SepaCVAE on DailyDialog. On DSTC7-AVSD, PVHD has a advantage of 10.50\% on diversity and 73.00\% on relevance compared to SepaCVAE. Compared to transformer-based baselines, although PVHD is sub-optimal in some metrics, it enjoys the advantage in most metrics, especially diversity. In terms of fluency, PVHD is only 1.00\% lower than HVRNN  and is much better that other baselines on DailyDialog. However, the fluency of PVHD is 26.50\% lower compared with HVRNN and 8.00\% lower compared with VHCR on DSTC7-AVSD. We argue that introducing a recurrent summary variable in the decoder increases the randomness of word generation, which will promote the diversity of the responses with a side effect of fluency reduction.
\subsection{Effectiveness of Summarizing  Variables}
We further analyze the effectiveness of PVHD on summarizing variables. Figure~\ref{visiual} demonstrates the visualization of word-level and utterance-level summarizing variables on on test set of DailyDialog and DSTC7-AVSD datasets. We can observe that both datasets exhibit high variability characteristic on word-level and utterance-level. Specifically, the summarizing variables on word-level show obvious categorical features, which indicates that a subsequence may have multiple suitable candidate words. Moreover, the summarizing variables on utterance-level also exhibit impressive categorical features, which confirms that there is a \textit{one-to-many} issue in the dialogue. 
These phenomena make dialogue generation different from machine translation where unique semantic mapping exists between source and target language.
\section{Conclusion}
We analyze the reasons for one-to-many and many-to-one issues from high variability of dialogue. 
We build PVHD based on proposed PVGRU component to model the word-level and utterance-level variation in dialogue for generating relevant and diverse responses. The results demonstrate that PVHD even outperforms pre-trained language models on diversity metrics.
\section{Limitations}
\label{app:lim}
Although our work can effectively model the variability issue in dialogue, we acknowledge some limitations of our study. Firstly, our study can work well on the approaches based on RNN, but cannot be employed to sequence models based on Transformer, which limits the generality of our approach. Secondly, although our methods can improve the diversity and relevence of responses, there are still gaps in fluency compared with other baselines.
\bibliography{anthology,custom}

\appendix
\begin{table*}[t]
\small
\centering
\begin{tabular}{lccccccccc}
\toprule
  Models & PPL & BLEU-1&BLEU-2 & Rouge-L &Dist-1&Dist-2 & Embed A&Embed E& Embed G \\
  \midrule
  PVHD &111.31& 32.19&25.42& 35.28& 15.33&49.93&  81.77&83.89&63.84 \\
  -RE & 127.73& 29.81&23.01& 29.88& 15.67&49.87&  80.72&83.89&61.33 \\
  -CO & 126.91& 31.19&24.49& 33.27& 12.44&48.34&  81.05&83.56&61.52 \\ \hline
\midrule
  PVHD &110.25& 29.87&20.03& 31.87& 6.54&31.77& 81.87&86.68&61.91\\
  -RE & 115.78& 24.01&13.34& 24.33& 7.42&30.96&  79.06&85.23&58.03\\
  -CO & 119.23& 28.28&18.68& 30.78& 5.89&27.47&  80.36&86.88&61.06 \\
\bottomrule
\end{tabular}
\caption{Ablation experiments of the loss module on test set of DailyDialog (up) and DSTC7-AVSD (down).}
\label{tab:ablation_study}
\end{table*}
\begin{table*}[ht]
\footnotesize
\centering
\begin{tabular}{l}
\toprule
  \textbf{Context}: Jenny, what's wrong with you? Why do you keep weeping like that?  Mary told me that she had seen you \\ \qquad with John last night. I got to know the fact that you are playing the field. \\ 
  \midrule
  \textbf{Gold:} honey, cross my heart, i'v never looked at another woman since the first day i set \\ \qquad \qquad my eyes on you! believe me, that's the truth.\\
  \textbf{SVT:} Honey, cross my heart, I'v never looked at another woman since the truth. \\
  \textbf{GVT:}  I'v never looked at another woman since the first day UNK. \\
  \textbf{SepaCVAE:}  i know how you really well. \\
  \textbf{DialogVED:} i'm sorry, but i can't let you do that. \\
  \textbf{PVHD:}  actually, i'v \textbf{not} looked at another woman \textbf{at all} \textbf{because} the first day i set my eyes on you! believe me. \\ \hline
\midrule
  \textbf{Context}: he turns his cellphone light on to help him see to screw the knob back into the dresser drawer. \\ \qquad \qquad does he end up fixing it correctly? yes, he screws the knob back on correctly. does he take \\ \qquad \qquad anything our of the drawer? no he does not open the drawer, only fixes it.\\ \hline
  \textbf{Gold:} he interacts with the computers after fixing the knob he simply stands up and begins to leave the room. \\
  \textbf{SVT:}  he appears to be carrying something.\\
  \textbf{GVT:} no, he does not go to the computer. \\
  \textbf{SepaCVAE:}  no, he does not move from his computer. \\
  \textbf{DialogVED:} no, he does not touch the computer. \\
  \textbf{PVHD:} no, he does \textbf{not} interact with the computer \textbf{at all}.  \\
\bottomrule
\end{tabular}
\caption{Examples of responses generated by the baselines. \textbf{Gold} represents the standard response provided by the dataset. \textbf{UNK} stands for unknown token.}
\label{tab:case_study}
\end{table*}
\begin{table*}[!ht]
\centering
\begin{tabular}{lccccc}
\toprule
  Item & SepaCVAE & SVT & GVT & PLATO & DialogVED \\
\midrule
  p-value & 0.0319 & 0.0107 & 0.0093 & 0.0032 &0.0246  \\ \hline
\midrule
  p-value & 0.0064 & 0.0475 & 0.0465 & 0.0080 &0.0447 \\
\bottomrule
\end{tabular}
\caption{Results of significance test of PVHD compared to other baselines on DailyDialog (up) and DSTC7-AVSD (down).} 
\label{tab:t-test}
\end{table*}
\section{Appendix}
\label{sec:appendix}
\subsection{Datasets}
\label{app:datasets}
To evaluate the performance of our proposed method, comprehensive experiments have been carried out on two publicly available datasets.
\textbf{DailyDialog}~\cite{li2017dailydialog} is a high-quality multi-turn dialogue dataset about daily life,  which consists of 11,118 context-response pairs for training, 1,000 pairs for validation, and 1,000 pairs for testing. In the experiment we abbreviate it as Daily.
\textbf{DSTC7-AVSD}~\cite{alamri2019audio}, short for Audio Visual Scene-aware Dialog of the DSTC7 challenge, is a multi-turn dialogue dataset from social media, which consists of 76,590 context-response pairs for training, 17,870 pairs for validation, and 1,710 pairs for testing. DSTC7-AVSD provides two available options of knowledge utilization: (i) textual knowledge including video’s caption and summary. (ii) multi-modal knowledge including text, audio and visual features. In this paper, we employ textual knowledge. In the experiment we abbreviate it as DSTC7.
\subsection{Implementation Details}
\label{app:impl}
We implement our model and baselines using Tensorflow 2 and train baselines on a server with RTX 8000 GPU (48G). The dimension of word embeddings is set 512. We consider at most 10 turns of dialogue context and 50 words for each utterance. The encoder adopts bidirectional structure and the decoder uses unidirectional structure. The hidden size of bidirectional encoder and bidirectional encoder is 1024 for VHCR, and other models are set 512. The size of latent variables for HVRNN, CVAE, VHCR, VAD, and SepaCVAE is 512. The size of summarizing variables for PVHD is 512. We set the number of encoder layers to 2 and the decoder layers to 1 for HVRNN, CVAE, VHCR, VAD, SepaCVAE and PVHD. The number of encoders and decoders are 4 for SVT and GVT. The head number of attention for SVT and GVT is 4. The batch size of VHCR is 32, and other models are 128. The init learning rate of HVRNN, CVAE, VAD, SepaCVAE, SVT, GVT and PVHD is set to 0.001. The learning rate of VHCR is set to 5e-4 and set to 3e-4 for DialogVED. We set the dropout rate of DialogVED to 0.1 and other baselines do not employ dropout trick. Adam~\cite{kingma2014adam} is utilized for optimization. The adam parameters beta1 and beta2 are set to 0.9 and 0.999, respectively. The maximum epoch is set to 100. Beam search is used to generate responses for evaluation. The beam size is set 5. The values of hyper-parameters described above are all fixed using the validation set.
\subsection{Automatic Evaluation Metrics}
\label{app:metrics}
We employ both automatic and human evaluations to assess the performance of compared methods. The automatic evaluation mainly includes the following metrics:
\textbf{BLEU}~\cite{yang2018adaptations} evaluates the n-gram co-occurrence between generated response and target response.
\textbf{ROUGE-L}~\cite{yang2018adaptations} evaluates the overlap of the longest common subsequences between generated response and the target response.
\textbf{Distinct-1/2}~\cite{li2015diversity} measures the generated response diversity, which is defined as the number of distinct uni-grams / bi-grams divided by the total amount of generated words.
\textbf{PPL} (Perplexity) evaluates the confidence of the generated response. The lower PPL score, the higher confidence for generating responses.
Embedding-based metrics (\textbf{Average, Exterma and Greedy)} measure the semantic relevance between generated response and target response~\cite{liu2016not,sedoc2019chateval,xu2018better}.
\subsection{Human Evaluation}
\label{app:human}
Following the work of~\cite{sun2021generating,li2017adversarial,xu2018diversity}, we divide six crowd-sourced graduate students into two groups to evaluate the quality of generated responses for 100 randomly sampled input contexts, respectively. 
We request annotators to rank the generated responses with respect to three aspects: fluency, diversity, and relevance. \textbf{Fluency} measures whether the generated responses are smooth or grammatically correct. \textbf{Diversity} evaluates whether the generated responses are informative, rather than generic and repeated information. \textbf{Relevance} evaluates whether the generated responses are relevant to the dialogue context. The average scores of the two groups is taken as the final score.
\subsection{Ablation Study}
\label{app:ablation}
We conduct ablation experiments on the proposed loss modules. Table~\ref{tab:ablation_study} reports the results of the ablation experiments of PVHD on DailyDialog and DSTC7-AVSD. \textbf{-RE} removes the reconstruction loss. \textbf{-CO} removes the consistency loss. The results demonstrate that our optimization objectives are effective. We can observe that the reconstruction loss can improve the BLEU-1/2 and Rouge-L. The consistency loss can improve Dist-1/2 metrics at the the expense of BLEU-1/2 and Rouge-L metrics. We believe that the consistency loss can ensure the consistency between the incremental information and the input at each time step. There may be multiple candidate tokens following the same distribution, which increases the diversity of generated responses. The reconstruction loss can make the summarizing variable recording the accumulated distribution of subsequence reflect the semantic information of dialogue context correctly, which will reduce the randomness of the generation process by limiting candidates that do not conform to sequence semantics.
\subsection{Significance Testing}
\label{app:sign}
To evaluate the reliability of the PVHD results, we performe multiple significance tests. Table~\ref{tab:t-test} (in Appendix~\ref{sec:appendix}) reports the results of the significance test for automatic evaluation. We can observe that the \textit{p-values} of PVHD are less than 0.05 compared with other models. Although the results of PVHD is not optimal in some metrics, the significance test demonstrates that results of PVHD are statistically significantly different from other models. In other words, the performance advantage of PVHD is statistically reliable and not an accident caused by random factors.
\subsection{Case Study}
\label{app:case}
To further dissect the quality of PVHD, several examples of generated responses are provided in Table~\ref{tab:case_study}. Although DialogVED, SVT, GVT can generate relevant responses, PVHD can produce higher quality responses in comparison. Specifically, for the first example, the responses generated by other models are contextual except for SepaCVAE.  The response generated by DialogVED is more diffuse than gold response, but response generated by PVHD is more informative and possesses a different sentence pattern and different wording than gold response to some extent. We can observe the similar case for the second example. We believe that this is mainly due to the capture of variability of corpus by summarizing variable, which enables the model to identify similar sentence patterns and words, and generate diverse responses.
\end{document}